\documentclass[10pt,twocolumn,letterpaper]{article}

\usepackage{cvpr}              

\usepackage[table]{xcolor}

\definecolor{cvprblue}{rgb}{0.21,0.49,0.74}
\usepackage[pagebackref,breaklinks,colorlinks,allcolors=cvprblue]{hyperref}

\def\paperID{10139} 
\def\confName{CVPR}
\def\confYear{2026}

\title{SpHOR: A Representation Learning Perspective on Open-set Recognition for Identifying Unknown Classes in Deep Neural Networks}

\author{Thiru Thillai Nadarasar Bahavan, Sachith Seneviratne, Saman Halgamuge\\
The University of Melbourne\\
Parksville\\
{\tt\small bahavant@student.unimelb.edu.au, \{sachith.seneviratne , saman.halgamuge\}@unimelb.edu.au}
}

\begin{document}
\maketitle
\begin{abstract}
The reliance on Deep Neural Network (DNN)-based classifiers in safety-critical and real-world applications necessitates Open-Set Recognition (OSR). OSR enables the identification of input data from classes unknown during training as unknown, as opposed to misclassifying them as belonging to a known class. DNNs consist of a feature extraction backbone and classifier head; however, most OSR methods typically train both components jointly, often yielding feature representations that adapt poorly to unknown data. Other approaches employ off-the-shelf objectives, such as supervised contrastive learning, which are not specifically designed for OSR. To address these limitations, we propose SpHOR, which explicitly shapes the feature space via supervised representation learning, before training a classifier. Instead of relying on generic feature learning, SpHOR custom-designs representation learning for OSR through three key innovations: (1) enforcing discriminative class-specific features via orthogonal label embeddings, ensuring clearer separation between classes. (2) imposing a spherical constraint, modeling representations as a mixture of von Mises-Fisher distributions. (3)  integrating Mixup and Label Smoothing (LS) directly into the representation learning stage. To quantify how these techniques enhance representations for OSR, we introduce two metrics: the Angular Separability (AS) and Norm Separability (NS). Combining all three innovations, SpHOR achieves state-of-the-art results (in AUROC and OSCR) across various coarse-grained and fine-grained open-set benchmarks, particularly excelling on the Semantic Shift Benchmark with improvements up to 5.1\%.

\end{abstract}

\section{Introduction}

\begin{figure}[htbp]
    \centering
    \includegraphics[width=0.9\linewidth]{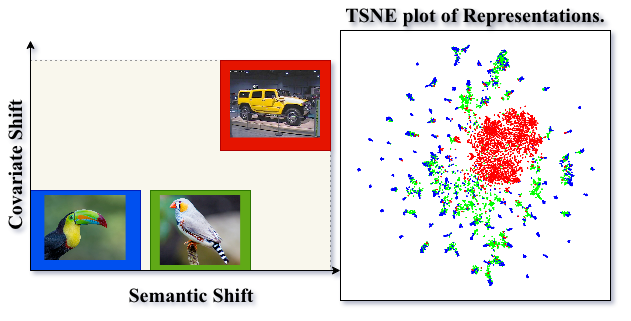}
    \caption{\textbf{The main challenge of OSR is that unknown-class test samples often lie very close to known training classes, making them difficult to separate.} The red points represent OOD samples that are easy to detect because they originate from a different data distribution (covariate shift) and exhibit a coarse-grained semantic shift. These samples are clearly separable from the blue samples (test samples from the known training classes). The green points represent unknown classes that share the same data distribution as the training set but differ semantically from known classes. Their close proximity to known-class samples makes them harder to detect, motivating our representation-specific approach, SpHOR, to handle such fine-grained novelties.}
    \label{fig:semvscov} 
\end{figure}

Commonly used machine learning algorithms for classification are closed-set. They assume that all classes associated with test data as known and already seen during training. However, in practical applications, the classes represented in training data can be incomplete, and unknown classes may be present in the test data. For example, a medical image classifier trained on five known skin cancer types may misclassify data belonging to a sixth type as one of the five. Open Set Recognition (OSR) addresses this issue by allowing systems to label samples from unknown classes as `unknown', while maintaining high accuracy on known classes \cite{Scheirer_2012}.

OSR and Out-of-Distribution (OOD) detection address different challenges. In OOD detection, test samples deviate from the training distribution due to covariate shifts, e.g., a known lesion in a different imaging modality. Semantic shifts (new categories) are secondary \cite{vaze2022openset,wang2024dissecting}. OSR, in contrast, focuses on semantic shifts, detecting previously unseen classes while maintaining accurate classification of known classes. As illustrated in Fig.~\ref{fig:semvscov}, this combination of requirements makes OSR particularly challenging.
 
This distinction has important implications for feature representation learning. In OSR, features must be class-specific. They should capture core characteristics rather than class-shared attributes like background patterns or texture \cite{vaze2022openset}. If the network focuses on class-shared attributes, unseen classes can be mapped closer to known classes in latent space. This reduces the effective semantic shift between known and unknown classes. This makes detecting unknown classes harder and can lead to the familiarity trap \cite{finegrainedosr, vaze2022openset}: a phenomenon where novel categories that are highly similar to the training categories are confused with familiar ones, often with high confidence, due to the reduced effective semantic shift. Closed-set classification assumes the entire feature space is occupied by known classes. In contrast, OSR explicitly reserves regions of the feature space for potential unknown classes, known as `open-space'\cite{scheirer2013open}. 

Deep neural networks (DNNs) typically consist of a feature representation extractor (implemented using CNNs, transformers, or similar architectures) followed by a classifier head~\cite{kang2019decoupling,XIONG2026112607,calibra}. In most OSR approaches, both components are trained simultaneously in one end-to-end process. This produces feature representations that are only implicitly adapted for unknown classes. The classifier primarily refines class boundaries rather than structuring the representation space, so open-space modelling emerges incidentally rather than as an explicit objective. Recently, Vaze~\cite{vaze2022openset} showed that even simple closed-set classification training strategies applied at the classifier level can outperform many advanced OSR methods, and that closed-set performance is strongly correlated with open-set performance. This raises a natural question: can explicitly designing the feature representations themselves, rather than relying solely on classifier-level training, further enhance open-set recognition? Decoupled training strategies, where representation extractor and classifier are trained separately, have shown success in long-tailed learning and calibration and offer a promising direction for explicitly shaping features in OSR\cite{kang2019decoupling,XIONG2026112607,calibra}. Some OSR methods employ general-purpose representation learning techniques, such as Supervised Contrastive Learning in ConOSR~\cite{conosr} in a decoupled scheme, but they are not explicitly designed to handle unknown classes. \textit{Thus, we are interested in the nature of OSR representations that lead to better open-set performance.}

Motivated by the challenge that generic feature learning is insufficient for detecting unknown classes, and building on recent insights~\cite{vaze2022openset,Dietterich2022FamiliarityHypothesis} and benchmarks, we introduce a representation learning method that enables improved feature representations for OSR through three key innovations: (1) Orthogonal separation of class-specific features and (2) Learning spherical representations that explicitly encourage Alignment and Uniformity in representations. (3) Integration of Mixup/LS into the representations.(Detailed in Section~\ref{prem})

In Summary, our contributions are: 
\begin{itemize}
\item We propose a novel \textbf{two-stage decoupled OSR training method}. Stage 1 learns class-specific representations using \textbf{orthogonal label embeddings} and model representations as a \textbf{mixture of von Mises-Fisher distributions}. In Stage 2, A classifier is fine-tuned on those representations.
\item We integrate Mixup and Label Smoothing into the representation learning stage, enhancing feature separability and robustness to unknown classes, specifically on reducing the `\textit{Familiarity Trap}'. To quantify their combined and individual effects, we introduce two evaluation metrics: \textbf{Angular Separability} and \textbf{Norm Separability}.
\item We analytically show how our loss induces \textbf{Alignment} and \textbf{Uniformity} in the representations.
\item  We achieve state-of-the-art results on various OSR/Semantic Shift benchmarks, with improvements up to \textbf{5.1\%} in OSCR and \textbf{5.2\%} in AUROC on the Semantic Shift Benchmark, which uses fine-grained datasets. 
\end{itemize}

\section{Related Work}


\textbf{Open-set Recognition}
Open-set recognition (OSR) methods are broadly categorized into discriminative and generative approaches~\cite{geng2020recent}. Discriminative approaches to open-set recognition (OSR) improve classifiers to recognize known classes while rejecting unknown ones, e.g., OpenMax~\cite{bendale2015towards} and PROSER~\cite{Zhou2021LearningPF}. Among discriminative approaches, prototype-based methods pursue a similar goal by representing classes through prototypes~\cite{Yang2020Convolutional, chen2020learning, chen2021adversarial}. However, both approaches struggle to capture the complexity of open-spaces and are limited by Euclidean representations~\cite{Wang2024ExploringDR} (Detailed in Sec~\ref{prem}). Generative methods model the data distribution or data manifold to detect unknowns, either by generating synthetic samples~\cite{Neal_2018_ECCV, kong2022opengan} or using reconstruction-based approaches~\cite{oza2019c2ae, huang2023class, sun2020open}. Hybrid methods combine both strategies~\cite{perera2020generative}. While effective, these approaches are computationally expensive, and can fail when unknown samples resemble known classes~\cite{conosr, vaze2022openset, huang2023class}. New approaches have begun exploring representation-learning-based methods for OSR \cite{conosr}.

According to Vaze et al.~\cite{vaze2022openset}, simple confidence-based baselines (e.g., those using logits) can match the performance of specialized OSR methods, suggesting that improvements in OSR often arise from advances in closed-set recognition enabled by more sophisticated architectures and training schemes. Building on their findings, we investigate, both theoretically and empirically, how specific training strategies enhance OSR performance via the lens of representation learning.

\textbf{Fine-grained OSR}
Conventional OSR methods perform well under coarse-grained semantic shifts but remain relatively unexplored in fine-grained settings. Recent work has begun addressing this gap~\cite{finegrainedosr, vaze2022openset}. Nico et al. propose a hierarchical adversarial learning scheme for fine-grained OSR~\cite{finegrainedosr}, but it relies on additional label granularity information, e.g., Taxonomy information,  limiting its applicability to broader domains.

\section{Preliminaries} \label{prem}
\noindent \textbf{Spherical Representations} Most OSR methods model representations in Euclidean space~\cite{chen2020learning, chen2021adversarial, Zhou2021LearningPF, Yang2020Convolutional}, where feature magnitudes can grow in an unrestricted way, leading to an unbounded open-space. This unboundedness significantly increases the open-space risk~\cite{scheirer2013open}, i.e., the likelihood of misclassifying known-class samples as unknown. The unbounded nature of Euclidean space exacerbates this risk, motivating strategies to constrain the open-space~\cite{Lubbering2022}. Thus, we constrain the open-space by L2-normalizing features, effectively projecting them onto a hypersphere. This allows us to model each class using a von Mises-Fisher distribution, which is naturally suited for spherical data. Such a formulation provides a mathematically interpretable and intuitive perspective on our method. 

\textit{The von Mises-Fisher (vMF) distribution} is the hyper-spherical analogue of the Gaussian distribution in Euclidean space\cite{MardiJupp1999, ming2024hypo}. The probability density function for a unit vector $\mathbf{z} \in \mathbb{R}^{p}$ in class $c$ is given by:
\begin{equation} \
    p(\mathbf{z} ; \boldsymbol{\mu}_{c}, \kappa) = R_{p}(\kappa) \exp \left(\kappa \boldsymbol{\mu}_{c}^\top \mathbf{z}\right),
\end{equation}
where $p$ is the dimensions of the projection, $\boldsymbol{\mu}_{c}$ is the normalized class label embedding on the unit-hyper-sphere, $\kappa$ is the concentration factor, A higher value of $\kappa$ results in a stronger density of the distribution around $\mu$. In the case when $\kappa$ approaches 0, the points become increasingly uniformly distributed across the hyper-sphere, and $R_{p}(\kappa)$ is a normalization factor~\cite{MardiJupp1999, ming2024hypo}. $\kappa$ can be interpreted as an inverse temperature $\frac{1}{\tau}$.

Crucially, compared to Euclidean representations, spherical representations allow us to study representations using the lens of alignment and uniformity \cite{wang2020understanding}. \textit{Alignment} is the expected pairwise distance between the positive example embeddings.  \textit{Uniformity} \cite{wang2020understanding} measures how uniformly spread out the embeddings in the representation space. Optimizing these properties as loss functions encourages the learned representations to exhibit a well-structured geometry, specifically improving linear separability between classes. \cite{wang2020understanding}. This is not always the case in standard Euclidean embedding spaces \cite{wang2020understanding}. Such linearly separated representations are particularly beneficial for open-set recognition, as they make it easier to identify samples that do not belong to any known class.

\noindent \textbf{Orthogonality Constraint} To address the `Familiarity Trap', it is essential to discourage features shared across classes in favor of class-specific features\cite{finegrainedosr}. To address this, our approach enforces subspace separation among class representations. Intuitively, this ensures that each class's feature vectors occupy distinct linear subspaces within the high-dimensional feature space, thereby encoding attributes specific to that class. Under this construction, Spaces that occupy multiple class-specific subspaces simultaneously or none at all will be assigned as the `open-space'. This orthogonality-based regularization offers several advantages over conventional margin-based methods, such as reduced reliance on arbitrary hyperparameters. Moreover, unlike margin-maximization or Equiangular Tight Frame (ETF)-based techniques \cite{MarkouNIPS2024}, it prevents negative correlations and feature redundancy that can arise when overlapping features are not explicitly controlled.

\section{Method}
\label{sec:method}

\begin{figure}[htbp]
    \centering
    \includegraphics[width=1.01\linewidth]{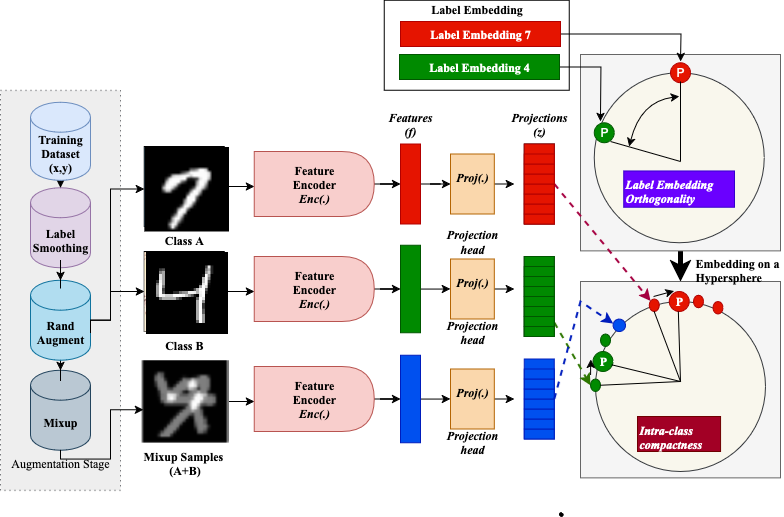}
    \caption{ \textbf{Data flow for Stage One of SpHOR.} The process begins with an augmentation stage, followed by the generation of feature projections. The loss function encourages each sample projection to align with its corresponding class label embedding, while simultaneously enforcing the label embeddings to be orthogonal to each other.
    }
    \label{fig:mainidea} 
\end{figure}

Stage One of our method focuses on representation learning (Section~\ref{stage1}), and Stage two trains the classifier using the learned representations (Section~\ref{stage2}). 

\subsection{Stage One: Spherical Representation Learning} \label{stage1}

We consider a batch of labeled training samples \(\{\boldsymbol{x}_i, \boldsymbol{y}_i\}_{i=1}^{N}\), where each \(\boldsymbol{x}_i\) is an image, and \(\boldsymbol{y}_i\) is its corresponding one-hot label selected from \( |C| \) classes. We apply RandAugment~\cite{cubuk2020randaugment} for image augmentation once per batch following the protocol in Vaze et~al. (\( \boldsymbol{x}_i \rightarrow \boldsymbol{x'}_i\))\cite{vaze2022openset}. Simultaneously, we apply label smoothing~\cite{muller2019when} to the one-hot labels (\( \boldsymbol{y}_i \rightarrow \boldsymbol{y'}_i\)). For the $i$-th training example, the smoothed label vector is defined as \( \boldsymbol{{y'}_i} = [y_{i1},y_{i2}, \dots, y_{iC}]\) where
\begin{equation}
y_{ij} = 1- \sigma, \quad y_{ik} = \frac{\sigma}{|C| - 1} \quad \text{for } k \neq j,
\end{equation}
 where $j$ is the index of the true class for training instance $i$, \( \sigma \) is the smoothing coefficient and \(|C| \) is the total number of classes. $y_{ij}$ refers to the smoothed probability assigned to the $j$-th class of the i-th training instance.

Afterwards, for each batch $\{\boldsymbol{x'}_i,\boldsymbol{y'}_i\}_{i=1}^{N}$, we apply Mixup~\cite{zhang2018mixup}:
\begin{equation}
\boldsymbol{\hat{x}}_{k} = \lambda \boldsymbol{{x'}}_{i} + (1 - \lambda) \boldsymbol{{x'}}_{j}, \quad \boldsymbol{\hat{y}}_{k} = \lambda \boldsymbol{{y'}}_{i} + (1 - \lambda) \boldsymbol{{y'}}_{j},
\end{equation}
where $\lambda$ is sampled from the Beta distribution, i.e. $\lambda \sim Beta(1,1)$. Here $i$ and $j$ are randomly sampled examples from the batch. All these steps are defined as the Augmentation Stage in Fig~\ref{fig:mainidea}. Then, $\{\boldsymbol{x'}_i,\boldsymbol{y'}_i\}_{i=1}^{N}$ and $\{\boldsymbol{\hat{x}}_i,\boldsymbol{\hat{y}}_i\}_{i=1}^{N}$ are then combined to form the unified augmented batch $\{\boldsymbol{\tilde{x}}_k, \boldsymbol{\tilde{y}}_k\}_{k=1}^{2N}$ for training. We will conduct ablation studies to investigate the effects of Mixup and Label Smoothing on OSR in further detail in Section~\ref{mixuplseffects}.


The unified augmented batch $\{\boldsymbol{\tilde{x}}_k, \boldsymbol{\tilde{y}}_k\}_{k=1}^{2N}$ is passed through an encoder network \( \text{Enc}(\cdot) \) (shown in Fig~\ref{fig:mainidea}), producing a $d$-dimensional embedding: \(\boldsymbol{f}_k~=~\text{Enc}(\boldsymbol{\tilde{x}}_k)\). These embeddings are further processed by a projection network, yielding $p$-dimensional projections \( \boldsymbol{\tilde{z}}_k~=~\text{Proj}(\boldsymbol{f}_k)\) and the projections are L2-normalized as \( \boldsymbol{z_{k}} = \boldsymbol{\tilde{z}_{k}} / \lVert \boldsymbol{\tilde{z}_{k}} \rVert_2 \). This follows the standard design established in supervised and self-supervised contrastive learning (e.g., SupCon and SimCLR)\cite{khosla2020supervised}. We use the von Mises-Fisher Alignment Loss to train the Feature Encoder, shown in Fig~\ref{fig:mainidea}, with normalized projections.


\noindent \textbf{von Mises-Fisher Alignment Loss (vMFAL)} \label{loss}

Under a mixture of von Mises-Fisher (vMF) distributions, one per class, an L2-normalized projection vector $\mathbf{z}_i$ is aligned with its corresponding L2-normalized label embedding (mean direction) $\boldsymbol{\mu}_c$ and $\kappa:=\frac{1}{\tau}$. The posterior probability of class $c$ is defined as


\begin{align}
 \label{prob}
    \mathbb{P}(y=c \mid \mathbf{z}_i; \{\boldsymbol{\mu}_{j}\}_{j=1}^C)
    &= \frac{R_{p}\left(\frac{1}{\tau}\right) 
        \exp \left(\frac{1}{\tau} \mathbf{z}_i^\top \boldsymbol{\mu}_{c}\right)}
        {\sum_{k \in C} R_{p}\left(\frac{1}{\tau}\right) 
        \exp \left(\frac{1}{\tau} \mathbf{z}_i^\top \boldsymbol{\mu}_{k}\right)} \\
    &= \frac{\exp \left(\mathbf{z}_i^\top \boldsymbol{\mu}_{c} / \tau\right)}
        {\sum_{k \in C} \exp \left(\mathbf{z}_i^\top \boldsymbol{\mu}_{k} / \tau\right)}.
\end{align}

The label embedding vectors $\boldsymbol{\mu}_{c}$ are initialized using Kaiming He initialization \cite{He_2015_ICCV} and always constrained to the unit sphere. We perform maximum likelihood estimation on Eq.~\ref{prob} with the training data to obtain:  

\begin{align}
 \label{spherical}
       \mathcal{L}_{\mathrm{vMFAL}}^{hard} :=- \frac{1}{N}\sum_{i=1}^N\log \frac{\exp \left(z_{i}^\top  {\boldsymbol{\mu}}_{c(i)} / \tau\right)}{\sum_{j \in C} \exp \left(z_{i}^\top  {\boldsymbol{\mu}}_{j} / \tau\right)},
\end{align}

where $i$ is the sample index, $c(i)$ is the target class of the $i^{th}$ sample and $N$ is the batch size.

However, $\mathcal{L}_{\mathrm{vMFAL}}^{hard}$ is only compatible with one-hot labels and is not compatible with Mixup or Label Smoothing. Therefore, we modify it to incorporate label similarity. Suppose we consider a sample $(x_{i}, y_{i})$, then our loss is: 

\begin{equation}
\mathcal{L}_{\mathrm{vMFAL}} 
:= - \frac{1}{N} \sum_{i=1}^N 
\underbrace{\sum_{k=1}^C S_{ik} \log P_{ik}}_{=: ~\mathcal{L}_{\mathrm{vMFAL}}^{(i)} \text{ for sample } i}
\end{equation}


Where  $P_{ik} = \frac{\exp(z_i^\top \boldsymbol{\mu}_k / \tau)}{\sum_{j=1}^C \exp(z_i^\top \boldsymbol{\mu}_j / \tau)}$ and $S_{ik} = \frac{y_{ik}}{\sum_{j=1}^C y_{ij}}$, respectively.

\noindent \textbf{Theorem 1}: \textit{The first derivative of the sample-wise loss $\mathcal{L}_{\text{\textit{vMFAL}}}^{(i)}$ for a sample \( i \) is:}

\begin{equation}
\frac{\partial \mathcal{L}_{\mathrm{vMFAL}}^{(i)}}{\partial z_i} 
  = -\sum_{k \in C} \left( S_{ik} - P_{ik} \right) \frac{{\boldsymbol{\mu}}_k}{\tau}
\end{equation}

Thus, minimizing this loss adjusts the model's parameter to align the label similarity \( S_{ik} \) with the class membership probability \( P_{ik} \). This loss structures the spherical representation space based on the label information. (Proof is in the Supplementary). The main advantage of this formulation is that samples with ambiguous semantics (mimicking unknown classes) can be generated via Mixup, thus leading to better modelling of the non-class specific spaces/open-space, leading to better OSR performance.

\noindent \textbf{Theorem 2}: \textit{$\mathcal{L}_{\text{\textit{vMFAL}}}$ promotes \textbf{Uniformity} and \textbf{Alignment} in the representations.}

\begin{equation}
\mathcal{L}_{\mathrm{vMFAL}}^{(i)} =
\underbrace{
-\frac{1}{\tau}\sum_{k=1}^C S_{ik} (z_i^\top \boldsymbol{\mu}_k)
}_{\text{Alignment}}
+
\underbrace{
\log \sum_{k=1}^C \exp\!\left(\frac{z_i^\top \boldsymbol{\mu}_k}{\tau}\right)
}_{\text{Uniformity}}
\end{equation}

\noindent The proposed $\mathcal{L}_{\mathrm{vMFAL}}$ loss effectively promotes \textbf{Alignment} by pulling feature representations ($\mathbf{z}_i$) toward their corresponding label embeddings ($\boldsymbol{\mu}_k$), and enforces \textbf{Uniformity} among the feature vectors by spreading them around the label prototypes~\cite{wang2020understanding}. If $\max(S_{ik}) \rightarrow 1$, then the alignment loss pulls the feature representation strongly towards the corresponding label embedding, dominating the loss over uniformity. However, if $\max(S_{ik}) \rightarrow \frac{1}{|C|}$, indicating an ambiguous sample, the feature representation is effectively pulled towards the mean of all label embeddings, $\frac{1}{|C|}\sum_{k=1}^C \boldsymbol{\mu}_k$, whose magnitude is smaller than that of an individual unit label embedding. In such a case, the uniformity loss takes over and spreads these samples between the label embeddings. Effectively, this forces the model to push ambiguous samples away from the class centers, resolving the `Familiarity Trap'.

Crucially, however, this uniformity property does not govern the relationship between the label embeddings ($\boldsymbol{\mu}_k$). This loss can be minimized even if all $\boldsymbol{\mu}k$ are highly similar or co-linear (\textit{label embedding collapse}), which would severely inhibit the model’s ability to discriminate between classes despite having uniformly distributed features. We explicitly resolve this issue via the \textit{Orthogonality Regularizer ($\mathcal{R}_{\mathrm{Ortho}}$)}.


\noindent \textbf{Orthogonality Regularizer ($R_{Ortho}$)}
To encourage the learning of distinct class-specific subspaces, we introduce an orthogonality regularizer that forces the label embeddings to be orthogonal and also uniform. This is more numerically stable and compatible with rectangular matrices than singular vector decomposition-based methods\cite{metaledi,8237672}.
\begin{equation}
       \mathcal{R}_{\mathrm{Ortho}} =  \log  \frac{1}{|C|^2-|C|} \sum_{\substack{~j \neq i}} \exp \left({\frac{1}{\tau}(\boldsymbol{\mu}_{j}\cdot {\boldsymbol{\mu}_{i}})^2} \right) 
\end{equation}

Overall training loss function is $\mathcal{L} = \mathcal{L}_{\mathrm{vMFAL}} +  \mathcal{R}_{\mathrm{Ortho}}$.

\subsection{Stage Two: Classifier Training}\label{stage2}
Just as in Supervised Contrastive/Contrastive Learning\cite{khosla2020supervised,contra}, after Stage One, we discard the projection network and the label embeddings. We generate features $f_i$ via the same training dataset (with the same Rand-Augment setting as Stage One ) from the encoder. We pass the frozen \textit{unnormalized} features $f_i$ via a classifier $H(\cdot)$ to get logits. We train the classifier only using the standard cross-entropy loss with minimal compute overhead.

\subsection{OSR Scoring Rules (Postprocessor )}\label{inference}

To address the binary task of determining whether a test sample belongs to the known training classes $\{1, \dots, C\}$ or an unknown class, we apply a \textit{scoring rule} $S(\cdot)$, which serves as a post-processing mechanism for OSR~\cite{vaze2022openset, wang2024dissecting}. Scoring rules $S(\cdot)$ can be categorized by the type of input they use: \textit{Classifier Scores} (e.g., MSP, MaxLogit) rely on classifier outputs, \textit{Feature Scores} (e.g., KNN) use the feature representations directly, and \textit{Hybrid Scores} that combine both.

A test input $\mathbf{x}_i$ is classified as a \textit{known class} if the \textit{scoring rule} $S(Enc(x_i); H(Enc(x_i))) \ge \theta $, and as an \textit{unknown class} otherwise, where $Enc(x_i)$ and $H(Enc(x_i))$ denote the feature representation and logits, respectively, and $\theta$ is the decision threshold~\cite{vaze2022openset}.  To comprehensively cover a wide spectrum of OSR scoring rules, we evaluate four distinct scoring rules that operate in different spaces: (1) Classifier Scores: MaxLogit\cite{vaze2022openset}, PostMax\cite{postmax} (2) Feature Scores: KNN\cite{sun2022knnood}, (3) Hybrid Scores: NNGuide\cite{10377411}.



\section{Experimental Setup} \label{experimentsssection}

Historically, most OSR research has relied on small-scale benchmarks (We refer to them as the Legacy CNN-32 OSR benchmark) that often fail to capture semantic shift, a key factor distinguishing OSR from general OOD detection~\cite{vaze2022openset, wang2024dissecting}. To address this, the Semantic Shift Benchmark (SSB) was proposed, highlighting varying levels of semantic shift (easy vs. hard) in fine-grained settings~\cite{wang2024dissecting, vaze2022openset, chen2021adversarial, Wang2024ExploringDR}. For coarser-grained shifts, which may also involve covariate shifts, we report results on legacy CNN-32 OSR benchmarks for the sake of completeness. All experiments were done on a 40GB Nvidia A100 GPU.

\subsection{Semantic Shift Benchmark} \label{ssbbench}

\textbf{Datasets} The benchmark uses three fine-grained classification datasets: CUB \cite{Wah2011CaltechUCSD}, FGVC-Aircraft \cite{Krause20133D}, and Stanford Cars \cite{Maji2013FineGrained}. Unknown classes are split into `Easy' and `Hard' sets based on semantic similarity~\cite{vaze2022openset}.

\textbf{Training Details} Due to the unavailability of the specific Places MocoV2 pretrained ResNet50 model referenced by Vaze et al. \cite{vaze2022openset}, we evaluated both ImageNet-pretrained and randomly initialized ResNet50 backbones to examine performance with and without standard pretraining. The baseline methods MLS \cite{vaze2022openset} and ARPL \cite{chen2021adversarial} were re-evaluated using the original hyperparameters reported by Vaze et al.\footnote{Since this benchmark does not provide a validation set, hyperparameters were not tuned on the test data; following prior work ensures a fair comparison.} ARPL/MLS both use label smoothing, for fairness, MLS was extended with Mixup as MLS+Mixup in benchmarking. SupCon and SpHOR share some hyperparameters with MLS, which we use whenever applicable. The learning rate is an exception and is taken from SupCon to accommodate differences in architecture. For hyperparameters not present in MLS, SpHOR inherits the corresponding SupCon values, reflecting the architectural similarities between the two methods. Both SupCon and SpHOR use a 1024-dimensional linear projection network in stage one and a linear classifier in stage two. All hyperparameters are reported in the Supplementary Material.

\textbf{Metrics} Similar to Vaze et al.~\cite{vaze2022openset} and report Top-1 multi-way classification accuracy (\textit{Acc.~\%}) to evaluate closed-set performance. Closed-set classification accuracy measures the proportion of correctly classified samples when the test set contains only known classes. For open-set evaluation, we report AUROC, which measures the area under the Receiver Operating Characteristic (ROC) curve in a threshold-independent manner and quantifies how well the model separates unknown-class samples from known-class samples. Note that AUROC does not consider classification accuracy. To jointly evaluate open-set detection and classification, we use the Open Set Classification Rate (OSCR) metric, which combines AUROC and classification accuracy across varying confidence thresholds \cite{OSCR}.

\begin{table*}[ht]
\centering
\scriptsize
\caption{OSR results on the \textbf{Semantic Shift Benchmark}, on the ResNet50 backbone with/without Imagenet-pretraining. Each method (except ARPL+) is paired with a scoring rule (MaxLogit/KNN/PostMax/NNGuide) for unknown detection. The table reports \textit{Closed-set Accuracy}, \textit{AUROC}, and \textit{OSCR}. Results are presented for both `Easy/Hard' unknown test-data splits. Best results in bold, second best underlined. The results were average over 3 different seeds. (w/o $R_{Ortho}$) denotes training without the orthogonality regularization term}
\begin{tabular}{l|ccc|ccc|ccc}
\hline
Semantic Shift & \multicolumn{3}{c}{\textbf{Caltech-UCSD-Birds (CUB)\cite{Wah2011CaltechUCSD}}} & \multicolumn{3}{c}{\textbf{Stanford Cars (SCars)\cite{Krause20133D}}} & \multicolumn{3}{c}{\textbf{FGVC-Aircraft\cite{Maji2013FineGrained}}} \\
Benchmark & Acc. ↑ & AUROC ↑  & OSCR ↑ & Acc. ↑ & AUROC ↑ & OSCR ↑ & Acc. ↑ & AUROC ↑ & OSCR ↑ \\
\rowcolor{blue!10}
With Imagenet Pretraining  & & (Easy/Hard) & (Easy/Hard) &  & (Easy/Hard) & (Easy/Hard) &  & (Easy/Hard) & (Easy/Hard)  \\ 

ARPL+~\cite{chen2021adversarial} & 85.4 & 81.77 / 73.88 & 73.07 / 66.94 & 89.8 & 84.95 / 76.39 & 79.32 / 72.21 & 83.3 & 85.75 / 74.57 & 75.51 / 66.69 \\
MLS~\cite{vaze2022openset} +MaxLogit & 85.2 & 83.16 / 73.98 & 74.38 / 67.16 & 89.3 & 83.96 / 76.63 & 78.32 / 72.33 & 80.2 & 85.23 / 73.73 & 73.01 / 73.01  \\
MLS~\cite{vaze2022openset} + Mixup +MaxLogit & 88.3 & 86.19 / 78.00 & 78.64 / 72.11 & 91.4 & 87.26 / 82.35 & 81.93 / 78.06 &  81.3 & 87.34 / 75.34 & 75.02 / 74.34  \\

MLS~\cite{vaze2022openset} + Mixup +PostMax & 88.3 & 84.34 / 74.44 & 77.03 / 68.82  &  91.4 & 85.75 / 81.32 & 81.25 / 77.75 &   81.3 &     81.24 / 71.14 & 71.12 / 71.22  \\
MLS~\cite{vaze2022openset} + Mixup +NNGuide & 88.3 & 86.56 / 77.14 & 77.88 / 70.30 & 91.5 & 76.34 / 76.65 & 71.86 / 72.26 &  81.3 &    79.34 / 78.45 & 69.32 / 68.78   \\
MLS~\cite{vaze2022openset} + Mixup +KNN & 88.3 & 81.43 / 69.62 & 73.33 / 63.33 & 91.5 & 72.84 / 73.23 & 68.58 / 68.92 &  81.3 &        78.99 / 77.89 & 67.31 / 66.11    \\
SupCON~\cite{khosla2020supervised} +MaxLogit & 78.2 & 88.52 / 75.44 & 72.62 / 63.55 & 91.8 & 90.90 / 79.25 & 85.61 / 75.42 & 88.9 & 83.02 / 79.93 & 77.79 / 75.36 \\
SupCON~\cite{khosla2020supervised} +PostMax & 78.2 & 84.09 / 68.32 & 67.91 / 56.58 & 91.8 & 82.53 / 60.17 & 77.55 / 56.94 & 88.9 & 64.84 / 58.95 & 60.67 / 55.47\\
SupCON~\cite{khosla2020supervised} +NNGuide & 78.2 & 80.73 / 67.25 & 66.67 / 56.43 &  91.8 & 86.81 / 77.67 & 81.46 / 73.43 & 88.9 & 56.25 / 65.58 & 52.74 / 61.30 \\
SupCON~\cite{khosla2020supervised} +KNN & 78.2 & 88.61 / 75.32 & 72.77 / 63.09 &  91.8 & 92.07 / 81.17 & 86.69 / 77.13 & 88.9 & 89.90 / 81.39 & 83.71 / 76.29 \\
\hline
SpHOR(w/o $R_{Ortho}$) +MaxLogit & \underline{90.3} & 91.10 / \textbf{83.59} & 85.09 / \bf{79.10} 

& \textbf{96.6} & \textbf{94.62} / \textbf{83.31} & \textbf{92.51} / \textbf{82.00} & 89.4 & 90.62 / \underline{82.26} & 85.00 / \underline{78.07} \\
SpHOR +MaxLogit & \textbf{90.8} & \textbf{91.72} / \underline{83.28} & \textbf{85.72} / \underline{79.01} 

& \underline{96.3} & \underline{94.12} / \underline{83.13} & \underline{91.76} / \underline{81.72} & 90.6 & 91.54 / 81.08 & 86.79 / 77.64 \\
SpHOR +KNN & 90.8 & 90.01 / 83.02 & 84.28 / 78.75 &  96.3 & 93.64 / 82.67 & 91.44 / 81.27 & 90.6 & 91.60 / 81.37 & 85.75 / 77.00 \\
SpHOR +NNGuide & 90.8 & 89.02 / 82.23 & 83.14 / 77.72 &  96.3 & 90.91 / 80.35 & 88.65 / 78.82 & 90.6 & \textbf{93.11} / \textbf{83.08} & \textbf{87.82} / \textbf{79.25} \\
SpHOR +PostMax & 90.8 & \underline{91.35} / 81.83 & \underline{85.35} / 77.60 & 96.3 & 93.44 / 82.57 & 91.12 / 81.17 & 90.6 & \underline{93.02} / 81.74 & 87.56 / 77.94\\

\hline
\rowcolor{blue!10}
Without Imagenet Pretraining & & (Easy/Hard) & (Easy/Hard) &  & (Easy/Hard) & (Easy/Hard) &  & (Easy/Hard) & (Easy/Hard)  \\ 

ARPL+~\cite{chen2021adversarial} & 46.8 & 67.59 / 59.35 & 37.33 / 33.95  & 76.1 & 80.45 / 72.38 & 66.70 / 61.18 &   77.8 & 78.53 / 70.18 & 67.33 / 61.41   \\ 
MLS~\cite{vaze2022openset} +MaxLogit &  53.2 & 68.42 / 61.79 & 43.44 / 41.26  &  76.5 & 79.31 / 71.22 & 65.12 / 60.34   &    78.2 & 78.66 / 72.23 & 67.45 / 62.89 \\
MLS~\cite{vaze2022openset} + Mixup +MaxLogit & 56.9 & 70.22 / 63.69 & 45.74 / 42.57 &  86.9 & 87.38 / 77.83 & 79.54 / 71.91  & 82.1 & 84.44 / 78.39 & 74.48 / 70.23  \\
MLS~\cite{vaze2022openset} + Mixup +PostMax & 56.9 & 73.10 / 62.93 & 47.10 / 42.01  &  86.9 & 86.39 / 77.13 & 78.76 / 71.32  &  82.1 & 80.65 / 73.86 & 72.30 / 67.28 \\
MLS~\cite{vaze2022openset} + Mixup +NNGuide & 56.9 & 59.38 / 59.65 & 38.75 / 38.66 &  86.9 & 81.37 / 74.81 & 74.10 / 68.56 &  82.1 & 77.49 / 79.10 & 67.45 / 68.34 \\
MLS~\cite{vaze2022openset} + Mixup +KNN & 56.9 & 52.53 / 53.96 & 31.01 / 31.80 &   86.9 & 63.42 / 65.01 & 57.88 / 59.16  &  82.1 & 79.17 / 78.92 & 67.81 / 67.55  \\

SupCON~\cite{khosla2020supervised} +MaxLogit & 77.8 & 87.86 / 77.06 & 71.73 / 64.71 & 78.7 & 75.17 / 65.74 & 63.73 / 56.42 & 66.7 & 82.55 / 77.71 & 59.64 / 57.04  \\
SupCON~\cite{khosla2020supervised} +PostMax & 77.8 & 85.02 / 69.59 & 68.75 / 57.87  &  78.7 & 75.83 / 68.97 & 62.78 / 57.93  &  66.7 & 45.95 / 66.78 & 31.61 / 46.27  \\
SupCON~\cite{khosla2020supervised} +KNN & 77.8 & 88.22 / 76.89 & 72.25 / 64.40 & 78.7 & 81.86 / 71.14 & 68.68 / 60.71 &  66.7 & 84.39 / 78.89 & 60.84 / 58.10 \\
SupCON~\cite{khosla2020supervised} +NNGuide & 77.8 & 81.24 / 70.64 & 66.63 / 58.94  &  78.7 & 71.23 / 63.56 & 60.08 / 54.02  &  66.7 & 68.79 / 71.57 & 50.41 / 52.62 \\
\hline

SpHOR(w/o $R_{Ortho}$) +MaxLogit & \underline{80.5} & 86.93 / 75.19 & 74.49 / 66.59 & \textbf{94.6} & 92.77 / 80.78 & \underline{89.52} / 78.77 & \textbf{88.8} & \underline{89.72} / \textbf{82.55}& \textbf{83.76} / \textbf{78.03} \\

SpHOR +MaxLogit & \textbf{82.7}& \textbf{87.47} / 77.52 & \textbf{76.72} / \underline{70.02} & \underline{94.4} & \textbf{93.01} / \underline{81.08} & \textbf{89.60} / \underline{78.97} & \underline{88.3} & 89.67 / \underline{81.46}& 83.41 / \underline{76.93} \\

SpHOR +PostMax  &82.7 & \underline{87.30} / 76.76 & \underline{76.54} / 69.19  & 94.4 & \underline{92.78} / 80.59 & 89.43 / 78.52 & 88.3 & \textbf{89.89} / 79.25 & 83.64 / 74.86\\
SpHOR +NNGuide & 82.7 & 83.33 / \textbf{77.82} & 72.82 / 68.94 & 94.4 & 91.54 / 80.84 & 88.41 / 78.75 &   88.3 & 86.33 / 80.28 & 81.15 / 76.15 \\ 
SpHOR  +KNN & 82.7 & 84.96 / \underline{77.77} & 75.38 / \textbf{70.29} & 94.4 & 91.80 / \textbf{80.94} & 88.89 / \textbf{79.12} &  88.3 & 88.98 / 80.55 & 82.67 / 76.09 \\

\end{tabular}

\label{tab:new_benchmark_results}
\end{table*}

\subsection{Legacy CNN-32 OSR Benchmarks}

\textbf{Training Details} Our method was evaluated against state-of-the-art OSR approaches on legacy CNN-32 benchmarks for fair comparison\footnote{CNN-32 is a 32-layer CNN similar to VGGnet\cite{vaze2022openset}; results are from papers with public GitHub repositories.}. We followed MLS~\cite{conosr,vaze2022openset} for hyper-parameters and training for the first stage.  Our method uses a 128-dimensional linear projection network in Stage one and a 128-node MLP in Stage two.

\textbf{Coarse-grained OSR Benchmark A} \label{unknowndetectionA}

\begin{table}[t]
\centering
\scriptsize
\caption{\textit{AUROC}-based comparison of OSR methods with CNN-32 backbone, averaged over five trials per dataset on the Legacy Benchmark A~\cite{neal2018open}. Best results in bold, second best underlined. }
\begin{tabular}{l|c|c|c|c|c}
\hline
\textbf{Method} & \textbf{SVHN} & \textbf{CIF10} & \textbf{CIF+10} & \textbf{CIF+50} & \textbf{TIN}   \\
\hline
\rowcolor{blue!10}
\textit{Openness\%} & \textit{22.5} & \textit{22.5} & \textit{46.5} & \textit{72.7} & \textit{68.3} \\
MLS\cite{hendrycks2016a} & 88.6 & 67.7 & 81.6 & 80.5 & 57.7 \\
OpenMax\cite{bendale2016towards} & 89.4 & 69.5 & 81.7 & 79.6 & 57.6 \\
OSRCI\cite{neal2018open} & 91.0 & 69.9 & 83.8 & 82.7 & 58.6 \\
C2AE\cite{oza2019c2ae} & 89.2 & 71.1 & 81.0 & 80.3 & 58.1\\
GFROSR\cite{perera2020generative} & 93.5 & 83.1 & 91.5 & 91.3 & 64.7 \\
PROSER\cite{Zhou2021LearningPF} & 94.3 & 89.1 & 96.0 & 95.3 & 69.3  \\
ARPL\cite{chen2021adversarial} & 95.3 & 91.0 & 97.1 & 95.1 & 78.2  \\
DIAS\cite{moon2022difficulty} & 94.7 & 85.0 & 92.0 & 91.6 & 73.1  \\
MLS\cite{vaze2022openset} & 97.1 & 93.6 & 97.9 & 96.5 &  \underline{83.0}  \\
ConOSR\cite{conosr} & \bf 99.1 & \underline{94.2} & \underline{98.1} & \bf 97.3 & 80.9  \\
MEDAF\cite{Wang2024ExploringDR} & 95.7 & 86.0 & 96.0 & 95.5 & 80.0 \\
BackMixup\cite{10923742} & 97.0 & 91.3 & 93.4 & 92.3 & 80.4\\

SpHOR(w/o$R_{Ortho}$) & 98.9 & 94.2 & 98.0 & 96.9 & \underline{83.8} \\
SpHOR & \bf 99.1 & \bf 94.5 & \bf{98.2} & \underline{97.2} & \bf 84.1  \\

\end{tabular}

\label{tab:table_auroc_old}
\end{table}

\textit{Datasets and Metrics} Our evaluation follows \cite{Neal_2018_ECCV, yoshihashi2019classification, perera2020generative}, averaging results across five splits per dataset, each with $N_{train}$ known and $N_{test}$ unknown classes. Dataset difficulty is measured by openness, $Openness~=1-\sqrt{\frac{N_{train}}{N_{test}}}$ \cite{conosr}. Our experiments utilize standard datasets: SVHN, CIFAR10(C10), and Tiny-ImageNet(TIN). For broader evaluation, we also include the common CIFAR10-CIFAR100 combinations known as CIFAR+10(C+10) and CIFAR+50(C+50), where 4 CIFAR10 classes serve as knowns, and 10 or 50 CIFAR100 classes serve as unknowns~\cite{Neal_2018_ECCV}. This particular benchmark reports only AUROC.

\begin{table}[ht]
\centering
\scriptsize

\setlength{\tabcolsep}{3pt}
\caption{Evaluation on the Legacy Benchmark B~\cite{chen2021adversarial} for OSR. CIFAR-10 is the known class; SVHN and CIFAR-100 are unknown. Results use a CNN-32 backbone~\cite{Wang2024ExploringDR}.}
\begin{tabular}{lcc|cc}
\toprule
\textbf{Method} & \textbf{DTACC} ↑ & \textbf{AUROC} ↑ & \textbf{DTACC} ↑ & \textbf{AUROC} ↑ \\
\rowcolor{blue!10}
 & \multicolumn{2}{c|}{(CIFAR10→SVHN)} & \multicolumn{2}{c}{(CIFAR10→CIFAR100)} \\
\midrule
CE\cite{hendrycks2016a} & 86.4 & 90.6 & 79.8 & 86.3 \\
GCPL\cite{Yang_2018_CVPR} & 86.1 & 91.3 & 80.2 & 86.4 \\
RPL\cite{Yang2020Convolutional} & 87.1 & 92.0 & 80.6 & 87.1 \\
ARPL\cite{chen2021adversarial} & 91.6 & 96.6 & 83.4 & 90.3 \\
ARPL+CSI\cite{chen2021adversarial} & 91.6 & 96.6 & 83.4 & 90.3 \\
OpenGAN\cite{Neal_2018_ECCV} & 92.1 & 95.9 & 84.2 & 89.7 \\
CSSR\cite{huang2023class} & 95.7 & 99.1 & 83.8 & 92.1 \\
RCSSR\cite{huang2023class} & 95.7 & 99.1 & 85.3 & 92.3 \\
MEDAF\cite{Wang2024ExploringDR} & 95.3 & 99.1 & {85.4} & 92.5 \\
BackMixup\cite{10923742} & 88.5 & 94.1 & 84.9 & 91.3 \\

SpHOR(w/o $R_{Ortho}$) & \underline{97.5} & \underline{99.1} & \textbf{86.7} & \underline{92.8} \\

SpHOR & \textbf{97.6} & \textbf{99.6} & \underline{86.4} & \textbf{93.2} \\

\end{tabular}

 \label{tab:nearfarood}
\end{table}
\textbf{Coarse-grained OSR Benchmark B} \label{unknowndetectionB}

\textit{Datasets} Chen et al.\cite{chen2020learning} introduced a legacy benchmark using CIFAR-10 as known classes. 
CIFAR10 $\rightarrow$ SVHN captures both semantic shift and covariate shift, whereas CIFAR10 $\rightarrow$ CIFAR100 captures mainly semantic shift (CIFAR10/CIFAR100 are sourced from the same Tiny Images dataset\cite{krizhevsky2009learning}).

\textit{Metrics} In addition to AUROC, this benchmark utilizes the following metrics such as DTACC which measures the highest classification accuracy for known and unknown samples across all possible thresholds. Additionally, metrics such as AUOUT/AUIN are detailed in the supplementary material.

\section{Results and Discussion} \label{resultssection}
\subsection{Benchmarking Results}

\noindent \textbf{SpHOR achieves top performance on the SSB}
Table~\ref{tab:new_benchmark_results} reports OSR results on CUB, Stanford Cars, and FGVC-Aircraft under Easy and Hard unknown splits. SpHOR consistently achieves top or near-top performance across both closed-set accuracy and open-set metrics. Even under Hard splits, performance remains strong, confirming SpHOR’s robustness to semantic shifts. 

\noindent \textbf{Among MLS+Mixup, SupCON, and SpHOR, SpHOR is the least sensitive to the choice of scoring rule.} Sensitivity was measured as the standard deviation of OSCR values across the four scoring functions (MaxLogit, PostMax, NNGuide, KNN), averaged across datasets. SpHOR shows the lowest sensitivity (easy/hard: 0.99 / 0.51), followed by SupCON (5.70 / 3.40), and MLS+Mixup, which has the highest sensitivity (6.04 / 3.52). Among scoring functions, MaxLogit consistently delivers the highest or second-highest AUROC and OSCR across datasets and training conditions, suggesting it is a strong choice for OSR tasks. We use it as a baseline throughout the paper.

\noindent \textbf{SpHOR remains robust without pretraining.} While baseline methods such as Mixup+CE and SupCon show substantial AUROC drops (up to ~20–30\%) when trained from scratch, SpHOR and SpHOR (+$R_{Ortho}$) maintain competitive performance. This shows that the spherical representation enables stable and discriminative learning, even when starting from poor initialization.

\noindent \textbf{Spherical representation enhances OSR separability compared to Euclidean variants}
SpHOR variants outperform Euclidean baselines such as ARPL+ and CE+Mixup under identical training conditions. This indicates that spherical normalization fosters better linear separability and more reliable open-set detection.

\noindent \textbf{SpHOR is effective for coarse-grained OSR on legacy benchmarks.}
On Benchmark A (Table~\ref{tab:table_auroc_old}), SpHOR attains the top performance, with a 0.81\% improvement on average AUROC (93.9 $\rightarrow$ 94.6) over the second best performing method ConOSR. On Benchmark B (Table~\ref{tab:nearfarood}), SpHOR attains a top performance, with a 1\% improvement (93.1 $\rightarrow$ 94.0) on average AUROC over the second best performing method RCSSR. These results confirm SpHOR’s robustness and versatility across both fine- and coarse-grained OSR scenarios.

\subsection{Ablation Study} \label{analysis}

\begin{table}[ht]
\centering
\scriptsize
\caption{Ablation study on the effects of Label Smoothing (LS) and Mixup when applied to representation learning within the SpHOR method (With MaxLogit Scoring rule). Results are averaged across three datasets from the SSB benchmark (SCARS, Aircraft, and CUB). Closed-set Accuracy, AUROC, and OSCR are reported for both easy and hard splits. Per-dataset results are provided in the Supplementary Material. }
\begin{tabular}{cc|ccc}
\hline
\textbf{Mixup} & \textbf{LS}  & \textbf{Avg. Accuracy ↑} & \textbf{Avg. AUROC ↑} & \textbf{Avg. OSCR ↑} \\
\rowcolor{blue!10}
 Stg. one   & Stg. one & & (Easy / Hard) & (Easy / Hard)  \\

\hline
x & x & 89.56 & 86.94 / 79.42 & 81.72 / 75.26 \\
x & \checkmark  & 89.00 & 90.22 / 79.10 & 84.01 / 74.74 \\
\checkmark & x & 91.66 & 91.32 / 80.85 & 86.78 / 77.40 \\
\checkmark & \checkmark & \textbf{92.60} & \textbf{93.00} / \textbf{83.20} & \textbf{88.40} / \textbf{80.00} \\

\end{tabular}

\label{tab:mixup_ablation}
\end{table}

\noindent \textbf{Label Smoothing and Mixup Optimize Angular and Norm Separability to Improve Representations for OSR.} \label{mixuplseffects} First, we investigate the individual and combined effects of two regularization strategies—Label Smoothing (LS) and Mixup, when applied directly to the learned representations, on the performance of the proposed SpHOR framework. The effects of these regularization methods are evaluated on the SSB benchmark consisting of three datasets: SCARS, Aircraft, and CUB. The evaluation considers three metrics: Closed-set Accuracy (Acc.), AUROC (measuring open-set performance), and OSCR (capturing the joint closed-set classification and open-set detection), \textit{averaged} across the three datasets. The results are summarized in Table~\ref{tab:mixup_ablation}. Identical patterns are observed for each dataset individually, and the detailed per-dataset results are provided in the Supplementary Material.

When Mixup is applied (\textit{Mixup} \checkmark, \textit{LS} X $\rightarrow$ \textit{Mixup} \checkmark, \textit{LS} \checkmark), it consistently improves performance across all metrics, regardless of whether Label Smoothing (LS) is used. In contrast, Label Smoothing exhibits a different pattern. When applied on top of Mixup (\textit{Mixup} \checkmark, \textit{LS} \checkmark), it has a synergistic effect, providing a modest but consistent improvement over using Mixup alone (\textit{Mixup} \checkmark, \textit{LS} X). However, when applied in isolation (\textit{LS} \checkmark, \textit{Mixup} X), the results are mixed: closed-set accuracy slightly decreases, while open-set performance improves for easy unknown classes but declines marginally for hard unknown classes. To gain deeper insight into how these regularization techniques, particularly their combination, achieve these results, we introduce two new metrics: Angular Separability (AS) and Norm Separability (NS).

The \textbf{\textit{Angular Separability} (\textit{AS})}, defined in Equation~\ref{SSI}, measures on average, how closely unknown samples ($u \in D_{u}$) lie near their nearest known class samples in the testing dataset ($ v \in D_{k} $). Thus, \textit{AS} captures the geometric/representational aspect of the familiarity trap. Lower \textit{AS} values indicate greater separability between known and unknown classes.

\begin{equation} \label{SSI}
    AS(D_{k},D_{u}) = \frac{1}{\left|D_{u}\right|} \sum_{u \in D_{u}} \max_{v \in D_{k}} \left( \frac{\mathbf{u} \cdot \mathbf{v}}{\|\mathbf{u}\| \|\mathbf{v}\|}\right)
\end{equation}

\begin{equation} \label{NSI}
NS = \mathrm{A^{UROC}} \big( \{\|\mathbf{v}\| : \mathbf{v} \in D_k\}, \{\|\mathbf{u}\| : \mathbf{u} \in D_u\} \big)
\end{equation}

\textbf{\textit{Norm Separability} (NS)}, defined in Equation~\ref{NSI}, measures the AUROC between the feature norms of the known class test samples and those of the unknown class samples. Higher values indicate that unknown classes can be readily detected via a straightforward, label-agnostic threshold applied to their feature norms. During the classifier training phase, we utilize unnormalized features. This is because the feature norm $||\mathbf{z}_i||$ carries crucial information about the model's uncertainty regarding an input sample\cite{scott2021von, park2023understanding}\footnote{Ablated in the Supplementary material.} 

\begin{table}[ht]
\centering
\scriptsize
\caption{Ablation study reporting \textit{Angular Separability (AS)} and \textit{Norm Separability (NS)} across SSB datasets under different \textbf{Mixup and Label Smoothing (LS)} settings.}
\begin{tabular}{cc|cc|cc|cc}
\hline
Mixup & LS  & \multicolumn{2}{c|}{\textbf{CUB}} & \multicolumn{2}{c|}{\textbf{SCARS}} & \multicolumn{2}{c}{\textbf{Aircraft}} \\
\rowcolor{blue!10}
& & AS ↓ & NS ↑ & AS ↓ & NS ↑ & AS ↓ & NS ↑ \\
\hline
x~&~x           & 0.6869 & 71.30 & 0.7765 & 84.18 & 0.7504 & 71.06 \\
\checkmark~&~x  & 0.6340 & 65.37 & 0.6917 & 72.48 & 0.6717 & 55.80 \\
\hline
x~&~\checkmark  & 0.7457 & 79.14 & 0.7824 & 86.92 & 0.8044 & 82.92 \\
\checkmark~&~\checkmark & 0.7121 & 78.67 & 0.7268 & 90.43 & 0.7197 & 84.96 \\
\hline
\end{tabular}

\label{tab:combined_side_by_side}
\end{table}

As observed in Table~\ref{tab:combined_side_by_side}, when LS is applied to the stage one, NS improves;however, AS degrades by a smaller scale. When Mixup is applied to stage one, AS improves;however, NS degrades by a smaller scale. This shows a contradictory relationship between LS and Mixup on AS and NS. When both are applied, although they are conflicting, the net effect yields a significant improvement in NS across all datasets, along with an improvement in AS on SCARS and Aircraft. While AS slightly degrades on CUB, the overall strong performance indicates that the combined effect successfully enhances feature separability, leveraging the strengths of each technique to improve different facets of the feature space for better open-set performance.

\begin{table}[ht]
\centering
\scriptsize
\caption{Ablation study investigating the impact of \textbf{Orthogonal Regularization ($R_{Ortho}$)} on \textbf{Dispersion (D)} and \textbf{AUROC} across datasets. Applying $R_{Ortho}$ generally improved \textbf{AUROC}, indicating that Orthogonality benefits OSR performance, specifically in detecting unknowns. MaxLogit was used as scoring rule.}
\begin{tabular}{c|cc|cc}
\hline
\textbf{Dataset} & \multicolumn{2}{c|}{\textbf{Dispersion ↑}} & \multicolumn{2}{c}{\textbf{AUROC ↑}} \\
\rowcolor{blue!10}
 &  & $+R_{Ortho}$ &  & $+R_{Ortho}$ \\ 
\hline
CUB & 77.22 & \textbf{81.22} & 86.93 / 75.19 & \textbf{87.47 / 77.52} \\
SCARS & 81.22 & \textbf{82.36} & 92.77 / 80.78 & \textbf{93.01 / 81.00} \\
Aircraft & 78.11 & \textbf{80.27} & \textbf{89.72 / 82.55} & 89.67 / 81.46 \\
TIN & 74.33 & \textbf{83.28} & 83.80 & \textbf{84.10} \\
\hline
\end{tabular}

\label{tab:combined_transposed}
\end{table}

\noindent  \textbf{Orthogonality Regularizer ($R_{Ortho}$) Optimize Dispersion to Improve Representations for OSR.} To understand the underlying mechanism, we introduce the \textbf{Dispersion (D)} metric, which quantifies the average angular distance between mean class features in the test set.  It is calculated as: $\text{D} = \frac{1}{|C|(|C| - 1)} \sum_{i=1} \sum_{\substack{j \in C~j \ne i}} \cos^{-1}(\mu_i^\top \mu_j)$. Here, $\mu_i$ represents the normalized mean feature vector for class $i$ within the test distribution. Higher $D$ indicates a more uniform distribution of mean class features in feature space, increasing \textit{Uniformity}, which we hypothesize facilitates open-set recognition. The ablation results in Table~\ref{tab:combined_transposed} demonstrate that $R_{Ortho}$ leads to higher AUROC, reflecting improved unknown detection in 2 out of 3 SSB datasets.  In Table~\ref{tab:new_benchmark_results}, however, this improvement also highlights dataset-dependent bias and requires further investigation. It should be noted that the improvements are incremental in some cases, particularly when using an ImageNet-pretrained backbone.

\noindent \textbf{Superior Training Efficiency and Small-Batch Robustness of the SpHOR Method.} The SpHOR method offers a clear computational advantage over SupCON. Its training complexity is \textit{linear}, scaling as $\mathcal{O}(B \cdot C)$ (where $B$ is the batch size and $C$ is the number of classes). This is significantly faster than SupCon’s \textit{quadratic} $\mathcal{O}(B^2)$ complexity, which arises from pairwise comparisons. Crucially, unlike contrastive methods that often require $B \gg C$ for stable training~\cite{contra}, SpHOR's class-based loss avoids pairwise sample comparisons, enabling robust convergence with \textit{significantly smaller batches}. This makes it ideal for resource-limited environments and a large number of classes. To verify empirically, on the Aircraft dataset, SpHOR (without Mixup) consistently outperforms SupCon under identical conditions, especially as batch size decreases (SpHOR / SupCon OSCR scores for $B=$ 64, 32, 16: \textit{78.7 / 72.4}, \textit{81.4 / 71.4}, and \textit{81.8 / 62.9}, respectively).

\section{Conclusions}
This paper introduces SpHOR, an OSR method that models classes as mixtures of von Mises-Fisher distributions with orthogonal label embeddings. By integrating Mixup, SpHOR significantly reduces unknown class misclassification and achieves state-of-the-art OSR performance on the SSB Benchmark. Ablation studies and new metrics confirm enhanced OSR performance.  

\section{Acknowledgement}
NB acknowledges Melbourne Graduate Research Scholarship. We would like to thank Chathura Jayasankha, Jayanie Bogahawatte, Yu Xia and Nisal Ranasinghe for providing valuable feedback. This research was supported by The University of Melbourne’s Research Computing Services .

{
    \small
    \bibliographystyle{ieeenat_fullname}
    \bibliography{main}
}


\end{document}